\def\year{2016}
\documentclass[letterpaper]{article}
\usepackage{aaai}
\usepackage{times}
\usepackage{helvet}
\usepackage{courier}
\usepackage{comment} 
\usepackage{graphicx} 
\usepackage{epstopdf} 
\usepackage{amsmath} 
\usepackage{mathrsfs} 
\usepackage{scrextend} 
\usepackage{url} 
\usepackage{subcaption} 
\usepackage{epstopdf}
\usepackage{bm} 
\usepackage{amssymb} 
\usepackage{color}
\usepackage{flushend} 
\newcommand{\tanhmine}{\operatorname{tanh}}
\newcommand{\drop}{\mathring} 

\makeatletter
\def\hlinew#1{%
  \noalign{\ifnum0=`}\fi\hrule \@height #1 \futurelet
   \reserved@a\@xhline}
\makeatother

\newcommand{\tabincell}[2]{\begin{tabular}{@{}#1@{}}#2\end{tabular}} 
\begin{document}
\title{Co-occurrence Feature Learning for Skeleton based Action Recognition\\using Regularized Deep LSTM Networks}
\author{
	{Wentao Zhu{\small $~^{1}$}\thanks{This work was done when W. Zhu was an intern at Microsoft Research Asia.}, Cuiling Lan{\small $~^{2}$}, Junliang Xing{\small $~^{3}$}, Wenjun Zeng{\small $~^{2}$}}, Yanghao Li{\small $~^{4}$}, Li Shen{\small $~^{5}$}, Xiaohui Xie{\small $~^{1}$}  \\
	$^{1}$\	University of California, Irvine, USA ~~ $^{2}$\,Microsoft Research Asia, Beijing, China\\
	$^{3}$\,Institute of Automation, Chinese Academy of Sciences, Beijing, China \\
	$^{4}$\ Peking University, Beijing, China ~~
	$^{5}$\,University of Chinese Academy of Sciences, Beijing, China \\	
	wentaoz1@uci.edu,
	\{culan,wezeng\}@microsoft.com, 
	jlxing@nlpr.ia.ac.cn,\\  
	lyttonhao@pku.edu.cn, li.shen@vipl.ict.ac.cn, xhx@ics.uci.edu	
}
\maketitle
\begin{abstract}
\begin{quote}
Skeleton based action recognition distinguishes human actions using the trajectories of skeleton joints, which provide a very good representation for describing actions. Considering that recurrent neural networks (RNNs) with Long Short-Term Memory (LSTM) can learn feature representations and model long-term temporal dependencies automatically, we propose an end-to-end fully connected deep LSTM network for skeleton based action recognition. Inspired by the observation that the co-occurrences of the joints intrinsically characterize human actions, we take the skeleton as the input at each time slot and introduce a novel regularization scheme to learn the co-occurrence features of skeleton joints. To train the deep LSTM network effectively, we propose a new dropout algorithm which simultaneously operates on the gates, cells, and output responses of the LSTM neurons. Experimental results on three human action recognition datasets consistently demonstrate the effectiveness of the proposed model.
\end{quote}
\end{abstract}

\section{1 ~~Introduction}
\label{sec_intro}
Recognizing human actions has remained one of the most important and challenging tasks in computer vision. It facilitates a wide range of applications such as intelligent video surveillance, human-computer interaction, and video understanding \cite{IVC10SurveyAction,CVIU11SurveyAction}.

Traditional studies on action recognition mainly focus on recognizing actions from RGB videos recorded by 2D cameras \cite{CVIU11SurveyAction}. However, capturing human actions in the full 3D space in which they actually occur can provide more comprehensive information. Biological observations suggest that humans can recognize actions from just the motion of a few light displays attached to the human body \cite{PP73Perception}. Motion capture systems \cite{MoCap} extract 3D joint positions using markers and high precision camera arrays. Although slightly higher in price, such systems  provide highly accurate joint positions for skeletons. Recently, the Kinect device has gained much popularity thanks to its excellent accuracy in human body modeling and affordable price. The bundled SDK for Kinect v2 can directly generate accurate skeletons in real-time. Due to the prevalence of these devices, skeleton based representations of the human body and its temporal evolution has become an attractive option for action recognition.

In this paper, we focus on the problem of skeleton based action recognition. The key to this problem  lies mainly in two aspects. One is to design robust and discriminative features from the skeleton (and the corresponding RGBD images) for intra-frame content representation \cite{ToG05MotionRetrieval,CVPR12Actionlet,ICRA12ActivityDetRGBD,VCIR14EigenJoints,ICMEWHumanInteraction}. The other is to explore  temporal dependencies of the inter-frame content for action dynamics modeling, using hierarchical maximum entropy Markov model \cite{PAIR11ActivityRGBD}, hidden Markov model \cite{CVPR12HO3DJ} or Conditional Random Fields \cite{ICCV05CRF}. Inspired by the success of deep recurrent neural networks (RNNs) using the Long Short-Term Memory (LSTM) architecture for speech feature learning and time series modeling \cite{ICASSP13DRNN,Graves2005Phoneme}, we intend to build an effective action recognition model based on deep LSTM network.

To this end, we propose an end-to-end fully connected deep LSTM network to perform automatic feature learning and motion modeling (Fig. \ref{fig:RDLSTM}). The proposed network is constructed by inheriting many insights from recent successful networks \cite{Book12Alex,NIPS12AlexNet,CVPR15Googlenet,CVPR15HRNN} and is designed to robustly model complex relationships among different joints. The LSTM layers and feedforward layers are alternately deployed to construct a deep network to capture the motion information. To ensure the model learns effective features and motion dynamics, we enforce different types of strong regularization in different parts of the model, which effectively mitigates over-fitting.

Specifically, two types of regularizations are proposed. (i) For the fully connected layers, we introduce regularization to drive the model to learn co-occurrence features of the joints at different layers. (ii) For the LSTM neurons, we derive a new dropout and apply it to the LSTM neurons in the last LSTM layer, which helps the network to learn complex motion dynamics. With these forms of regularization, we validate our deep LSTM networks on three public datasets for human action recognition. The proposed model has been shown to consistently outperform other state-of-the-art algorithms for skeleton based human action recognition.

\section{2 ~~Related Work}
\label{sec:relatedwork}

\subsection{2.1 ~~Activity Recognition with Neural Networks}
In contrast to the handcrafted features, there is a growing trend of learning robust feature representations from raw data with deep neural networks, and excellent performance has been reported in image classification \cite{NIPS12AlexNet} and speech recognition \cite{ICASSP13DRNN}. However, there are only few works which leverage neural networks for skeleton based action recognition. A multi-layer perceptron network is trained to classify each frame \cite{VISAPP14Classify}; however, such a network cannot explore temporal dependencies very well. In contrast, a gesture recognition system \cite{Gesture:2013} employs a shallow bidirectional LSTM with only one forward hidden layer and one backward hidden layer to explore long-range temporal dependencies. A deep recurrent neural network architecture with handcrafted subnets is utilized for skeleton based action recognition \cite{CVPR15HRNN}. However, the handcrafted hierarchical subnets and their fusion ignore the inherent co-occurrences of joints. This motivates us to design a deep fully connected neural network which is capable of fully exploiting the inherent correlations among skeleton joints in various actions.

\subsection{2.2 ~~Co-occurrence Exploration} 
An action is usually only associated with and characterized by the interactions and combinations of a subset of the skeleton joints. For example, the joints ``hand", ``arm" and ``head" are associated with the action ``making telephone call". An actionlet ensemble model exploits this trait by mining some particular conjunctions of the features corresponded to some subsets of the joints \cite{CVPR12Actionlet}. Similarly, actions involving two people can be characterized by the  interactions of a subset of the two persons' joints \cite{CVPRW12TwoPerson,ICMEWHumanInteraction}. Inspired by the actionlet ensemble model, we introduce a new exploration mechanism in the deep LSTM architecture to achieve automatic co-occurrence mining as opposed to pre-specifying in advance which joints should be grouped.

\subsection{2.3 ~~Dropout for Recurrent Neural Networks}
Dropout has been demonstrated to be quite effective in deep convolutional neural networks \cite{NIPS12AlexNet}, but there has been relatively little research on applying it to RNNs. In order to preserve the ability of RNNs to model sequences, dropout applied only to the feedforward (along layers) connections but not to the recurrent (along time) connections is proposed \cite{ICFHR14DropoutRCNN}. This is to avoid erasing all the information from the units (due to dropout). Note that the previous work only considers dropout at the output response for an LSTM neuron \cite{ICLR15DropoutLSTM}. However, considering that an LSTM neuron consists of internal cell and gate units, we believe one should not only look at the output of the neuron but also into its internal structure to design effective dropout schemes. In this paper, we design an in-depth dropout for LSTM to address this problem.
\begin{figure}[t]
\centering\includegraphics[width=0.99\linewidth]{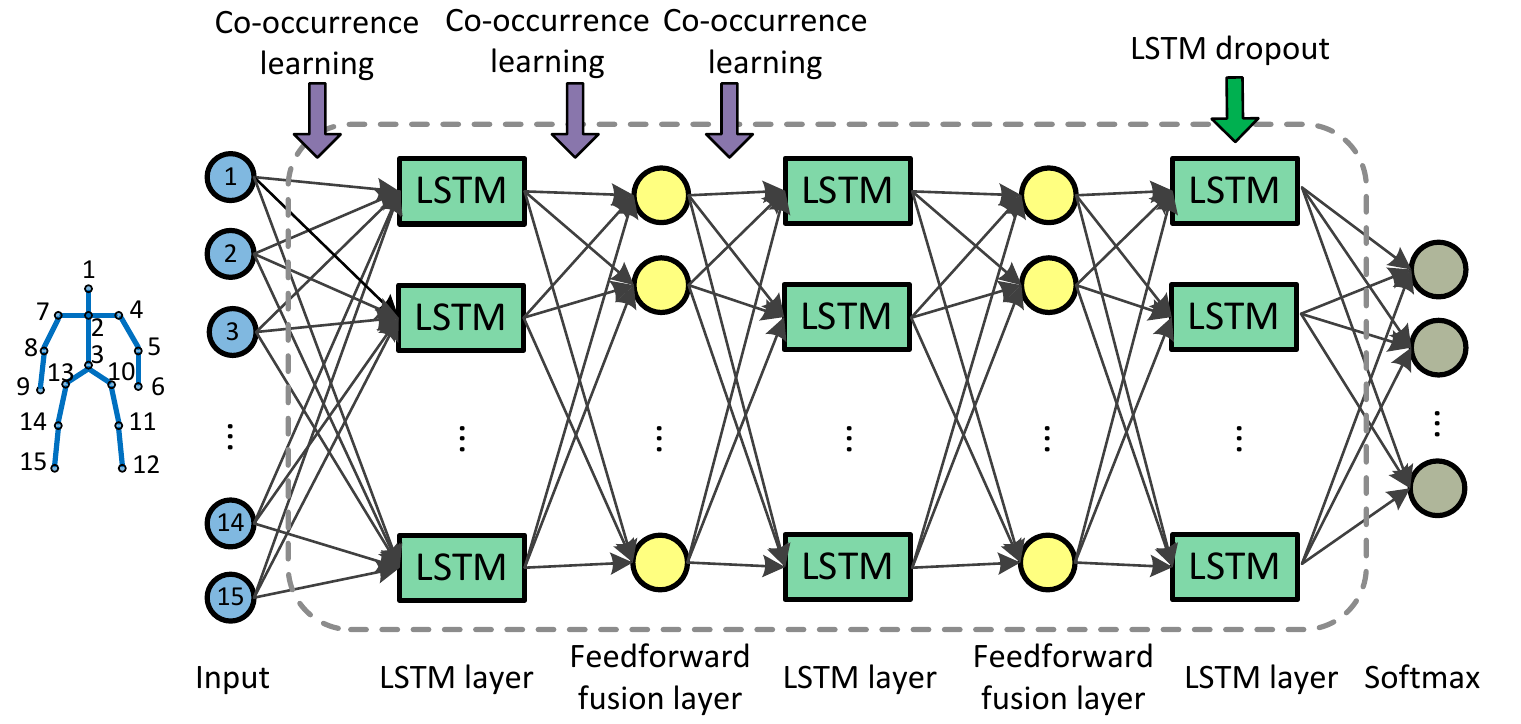}
\caption{The proposed deep LSTM network with three LSTM layers and two feedforward layers. For clarity, the temporal recurrent structure is not shown.}\label{fig:RDLSTM}
\end{figure}
\section{3 ~~Deep LSTM with  Co-occurrence Exploration and In-depth Dropout}
\label{sec_approach}
Leveraging the insights from recent successful networks, we design a fully connected deep LSTM network for skeleton based action recognition. Fig. \ref{fig:RDLSTM} shows the architecture of the proposed network, which has three bidirectional LSTM layers, two feedforward layers, and a softmax layer that gives the predictions. The proposed full connection architecture enables one to fully exploit the inherent correlations among skeleton joints. In the network, the co-occurrence exploration is applied to the connections prior to the second LSTM layer to learn the co-occurrences of joints/features. LSTM dropout is applied to the last LSTM layer to enable more effective learning. Note that each LSTM layer uses bidirectional LSTM and we do not explicitly distinguish the forward and backward LSTM neurons in Fig. \ref{fig:RDLSTM}. At each time step, the input to the network is a vector denoting the 3D positions of the skeleton joints in a frame. 

In the following, we first review LSTM briefly to make the paper self-contained. Then we introduce our method for co-occurrence exploration in the deep LSTM network. Lastly we describe our dropout algorithm which is designed for the LSTM neurons and enables effective learning of the model. 

\subsection{3.1 ~~Overview of LSTM}
The RNN is a successful model for sequential learning \cite{Book12Alex}. For the recurrent neurons at some layer, the output responses ${\bf{h}}_{t}$ are calculated based on the inputs ${\bf{x}}_{t}$ to this layer and the responses ${\bf{h}}_{t-1}$ from the previous time slot
\begin{equation}
\label{equ:rnn}
{{\bf{h}}_{t}} = \theta \left( {{\bf{W}}_{xh}}{{\bf{x}}_{t}} + {{\bf{W}}_{hh}}{{\bf{h}}_{t-1} + {\bf{b}}_h} \right),
\end{equation}
where $\theta \left(\cdot \right)$ denotes the activation function, ${\bf{b}}_h$ denotes the bias vector, ${{\bf{W}}_{xh}}$ is the matrix of weights between the input and hidden layer and ${\bf{W}}_{hh}$ is the matrix of recurrent weights from the hidden layer to itself at adjacent time steps which is used for exploring temporal dependency. 

\begin{figure}[t]
	\centering\includegraphics[width=0.65\linewidth]{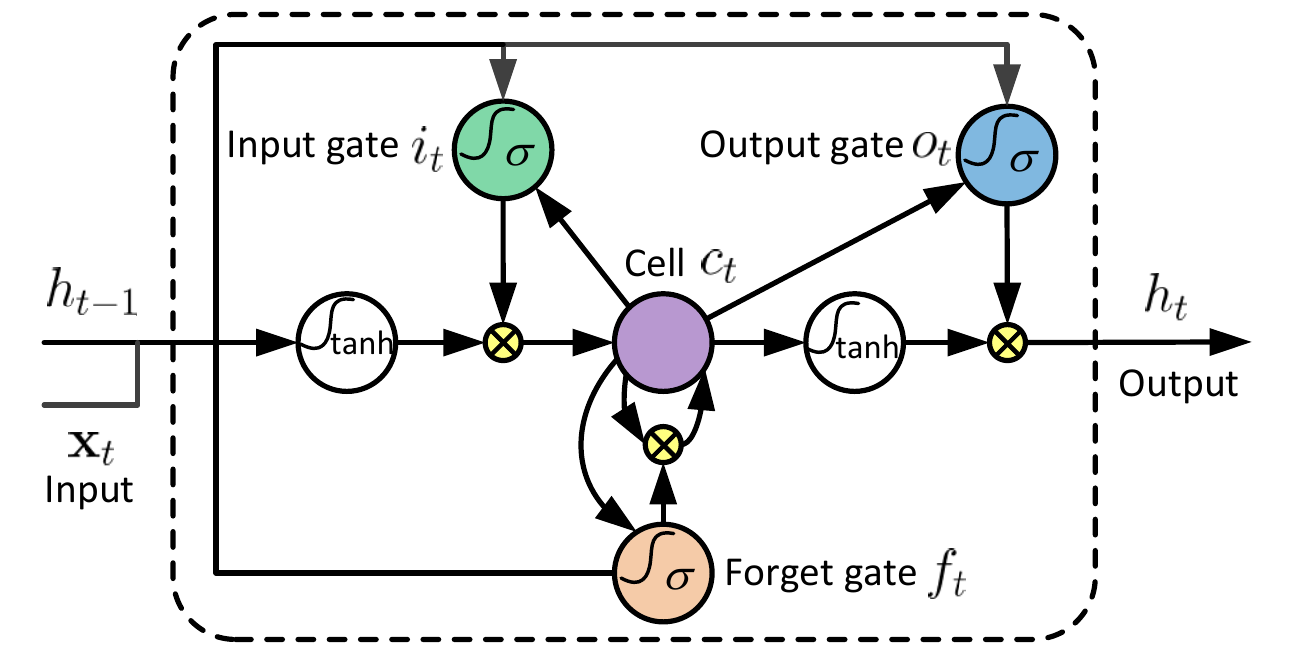} 
	\caption{The structure of an LSTM neuron.}\label{fig:lstm}
\end{figure}

LSTM is an advanced RNN architecture which can learn long-range dependencies \cite{ICASSP13DRNN}. Fig. \ref{fig:lstm} shows a typical LSTM neuron, which contains an input gate ${{i}}_t$, a forget gate ${{f}}_t$, a cell ${{c}}_t$, an output gate ${{o}}_t$ and an output response $h_t$. The input gate and forget gate govern the information flow into and out of the cell. The output gate controls how much information from the cell is passed to the output ${{h}}_t$. The memory cell has a self-connected recurrent edge of weight 1, ensuring that the gradient can pass across many time steps without vanishing or exploding. Therefore, it overcomes the difficulties in training the RNN model caused by the ``vanishing gradient'' effect. For all the LSTM neurons in some layer, at time $t$, the recursive computation of activations of the units is
\begin{eqnarray}
\label{equ:lstm}
\begin{aligned}
\!&{\bf{i}}_t = \sigma \left( {{\bf{W}}_{xi}}{{\bf{x}}_{t}} + {{\bf{W}}_{hi}}{\bf{h}}_{t-1} + {{\bf{W}}_{ci}}{\bf{c}}_{t-1} + {\bf{b}}_i \right), \\ 
\!&{\bf{f}}_t = \sigma \left( {{\bf{W}}_{xf}}{{\bf{x}}_{t}} + {{\bf{W}}_{hf}}{\bf{h}}_{t-1} + {{\bf{W}}_{cf}}{\bf{c}}_{t-1} + {\bf{b}}_f \right), \\
\!&{\bf{c}}_t = {\bf{f}}_t\!\odot {\bf{c}}_{t-1}\!+{\bf{i}}_t \odot \tanhmine\! \left( {{\bf{W}}_{xc}}{{\bf{x}}_{t}}\! +\! {{\bf{W}}_{hc}}{\bf{h}}_{t-1} \!+\! {\bf{b}}_c \right), \\ 
\!&{\bf{o}}_t = \sigma \left( {{\bf{W}}_{xo}}{{\bf{x}}_{t}} + {{\bf{W}}_{ho}}{\bf{h}}_{t-1} + {{\bf{W}}_{co}}{\bf{c}}_{t} + {\bf{b}}_o \right), \\
\!&{\bf{h}}_t = {\bf{o}}_t \odot \tanhmine \left( {\bf{c}}_t \right),
\end{aligned}
\end{eqnarray}
where $\odot$ denotes element-wise product, $\sigma \left(x \right)$ is the sigmoid function defined as $\sigma\left(x\right)=1/(1+e^{-x})$, ${\bf{W}}_{\alpha \beta}$ is the weight matrix between $\alpha$ and $\beta$ (e.g., ${\bf{W}}_{x i}$ is the weight matrix from the inputs ${\bf{x}}_{t}$ to the input gates ${\bf{i}}_t$), and ${\bf{b}}_{\beta}$ denotes the bias term of $\beta$ with $\beta \in \{i,f,c,o\}$. Four weight matrixes are associated with input ${\bf{x}}_{t}$. To allow the information from both the future and the past to determine the output, bidirectional LSTM can be utilized \cite{Book12Alex}.
\subsection{3.2 ~~Co-occurrence Exploration}
\label{subsec:co-occur}
The fully connected deep LSTM network in Fig. \ref{fig:RDLSTM} has very powerful learning capability. However, it is difficult to learn directly due to the huge parameter space. To overcome this problem, we introduce a co-occurrence exploration process to ensure the deep model learns effective features.
\begin{figure}[th]
	\centering
	\begin{subfigure}[t]{0.0485\textwidth}
		\centering\includegraphics[width=\textwidth]{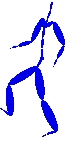} 
	\end{subfigure}
	~
	\begin{subfigure}[t]{0.166\textwidth}
		\centering\includegraphics[width=\textwidth]{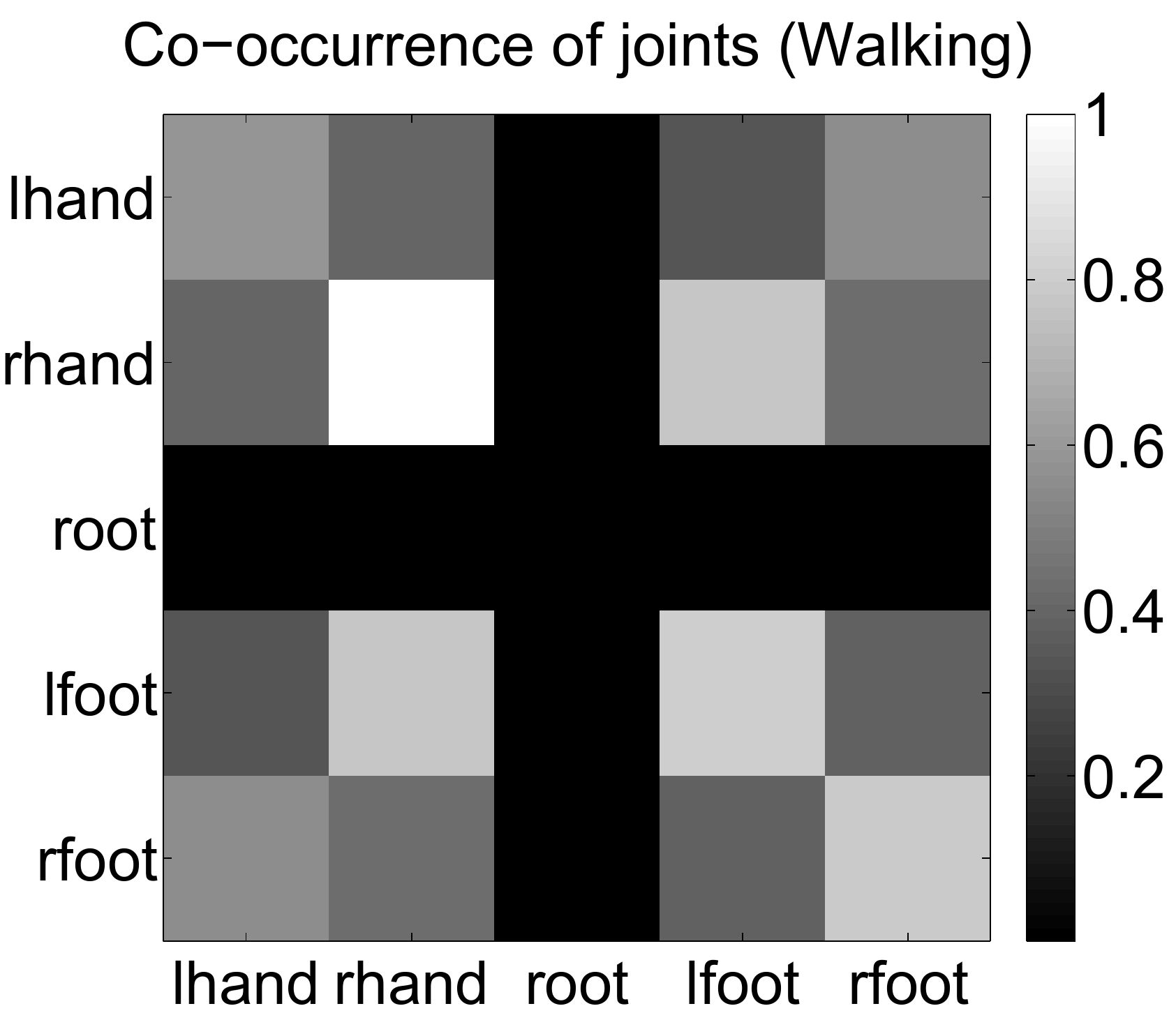}
	\end{subfigure}
	~
	\begin{subfigure}[t]{0.051\textwidth}
		\centering\includegraphics[width=\textwidth]{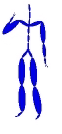}
	\end{subfigure}
	~
	\begin{subfigure}[t]{0.166\textwidth}
		\centering\includegraphics[width=\textwidth]{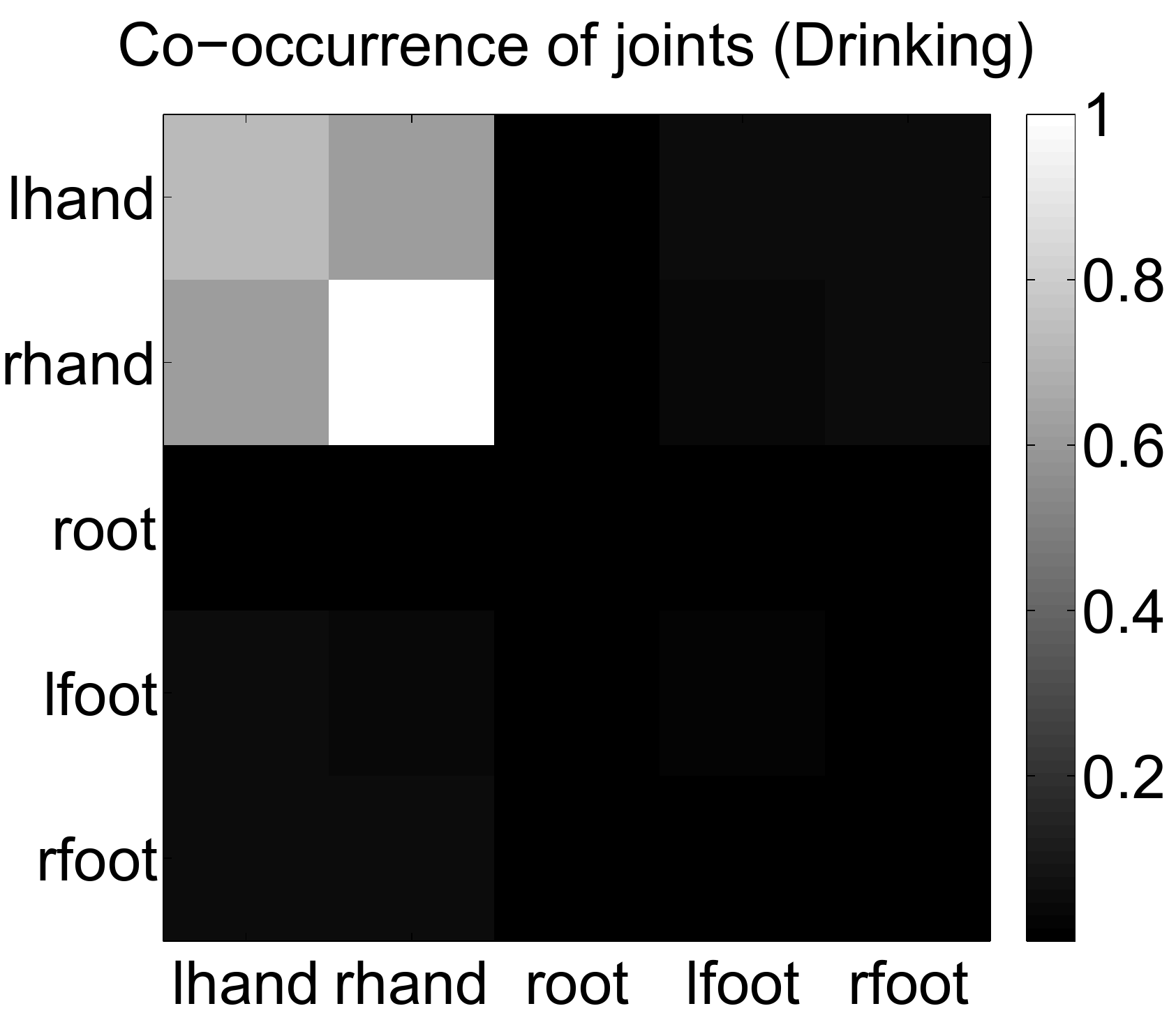}
	\end{subfigure}
	\caption[]{Illustration of co-occurrences of joints for ``walking" and ``drinking" respectively (using the absolute values of the covariance matrix). Joints from different parts are active simultaneously, e.g., joints of hands and feet for ``walking". Different actions have different active joint sets. }\label{fig:cooccurrencemotivation}
\end{figure}

\begin{figure}[th]
	\centering\includegraphics[width=0.99\linewidth]{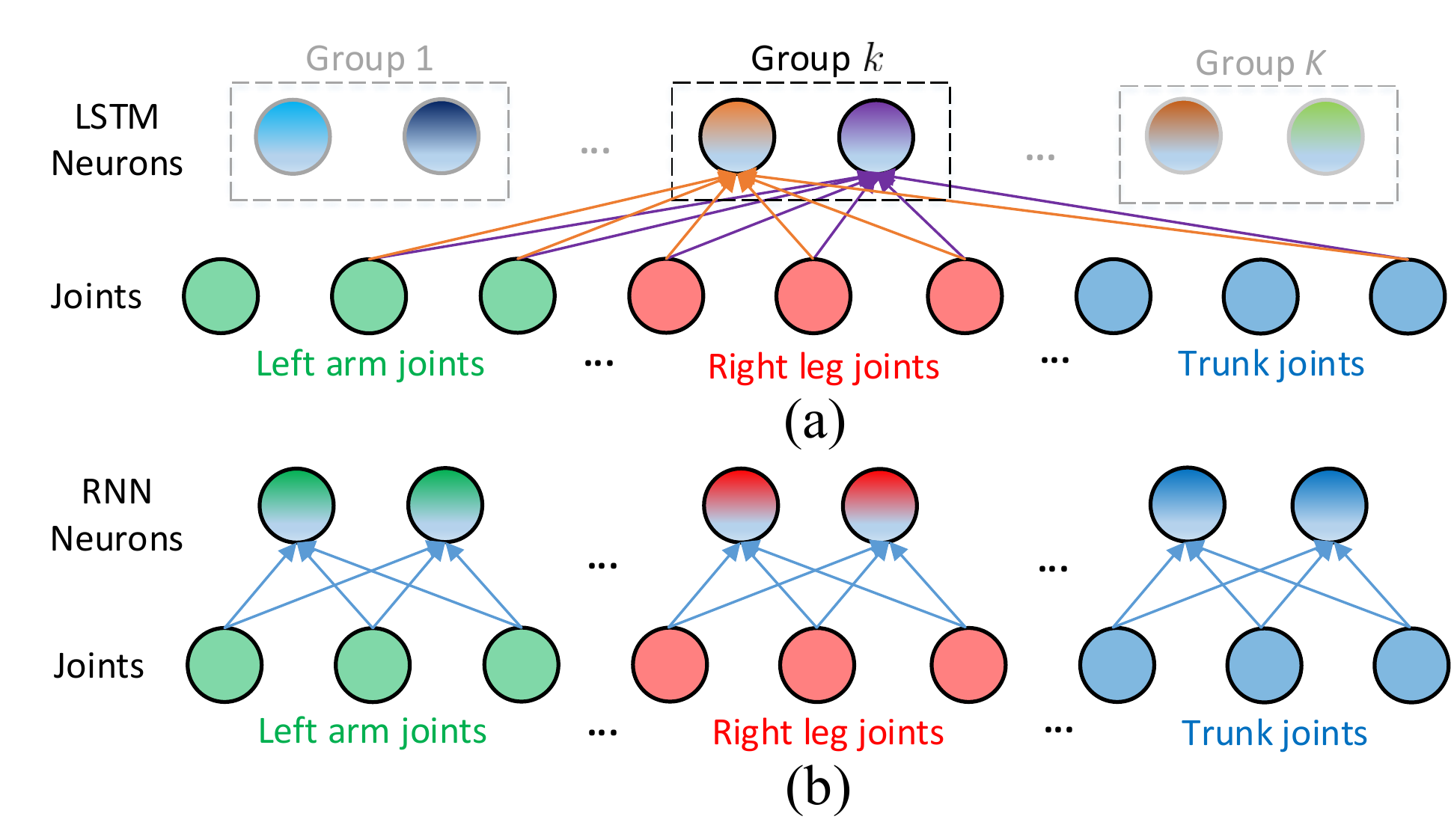}
	\caption{Illustration of the connections between joints and neurons in the first layer. (a) Co-occurrence connections automatically learned for Group $k$ (proposed). (b) Part-based subnet connections \cite{CVPR15HRNN}, where the co-occurrences of joints across different parts are prohibited.}\label{fig:Group}
\end{figure}

The co-occurrence of some joints can intrinsically characterize a human action. Fig. \ref{fig:cooccurrencemotivation} shows two examples. For ``walking", the joints from hands and feet have high correlations but they all have low correlations with the joint of root. The sets of correlated joints for ``walking" and ``drinking" are very different, indicating the discriminative subset of joints varies for different types of actions. Two main aspects have been considered in our design of the network and the specialized regularization we propose. (i) We expect the network can automatically explore the conjunctions of discriminative joints. (ii) We expect the network can explore different co-occurrences for different types of actions. Therefore, we design the fully connected network to allow each neuron being connected to any joints (for the first layer) or responses of the previous layer (for the second or higher layer) to automatically explore the co-occurrences. Note that the output responses are also referred to as \emph{features} which are the input of the next layer. We divide the neurons in the same layer into $K$ groups to allow different groups to focus on exploration of different conjunctions of discriminative joints. Taking the $k^{th}$ group of neurons as an example (see Fig. \ref{fig:Group} (a)), the neurons will automatically connect the discriminative joints/features. In our design, we incorporate the co-occurrence regularization into the loss function
\begin{equation}
\min_{{\bf{W}}_{x\beta}}{\mathcal{L}}\!+{\lambda}_1\!\!\sum_{\substack{\beta \in S}}{\left\Vert {\bf{W}}_{x\beta}\right\Vert}_{1}\!\!+\!{\lambda}_2\!\sum_{\substack{\beta \in S}}\sum_{k=1}^{K}  \left\Vert {{\bf{W}}_{x\beta,k}}^T \right\Vert _{2,1}\!,
\label{equ:opt}
\end{equation}
where ${\mathcal{L}}$ is the maximum likelihood loss function of the deep LSTM network \cite{Book12Alex}. The other two terms are used for the co-occurrence regularization which can be applied to each layer. ${\bf{W}}_{x\beta}\!=\![{\bf{W}}_{x\beta,1};\cdots;{\bf{W}}_{x\beta,K}]\!\in {\mathbb{R}}^{N \times J} $ is the connection weight matrix from inputs to the units associated with $\beta \in S$, with $N$ indicating the number of neurons and $J$ the dimension of inputs (e.g., for the first layer, $J$ is the number of joints multiplied by 3). The $N$ neurons are partitioned into $K$ groups and the number of neurons in a group is $L\!=\!\lceil N/K \rceil$, with $\lceil \cdot \rceil$ representing the rounding up operation. ${\bf{W}}_{x\beta,k}$ is the matrix composed of the $(L(k-1)\!+\!1)^{th}$ to $(Lk)^{th}$ rows of ${\bf{W}}_{x\beta}$. ${{\bf{W}}_{x\beta,k}}^T$ denotes its transpose. ${S}$ denotes the set of internal units which are directly connected to the input of a neuron. For the LSTM layer, $S\!=\!\{i, f, c, o\}$ denotes the gates and cell in LSTM neurons. For the feedforward layer, $S\!=\!\{h\}$ denotes the neuron itself. 

In the third term, for each group of units, a structural $\ell_{21}$ norm, which is defined as $\left\Vert {\bf{W}}\right\Vert_{2,1}\!\!=\!\!\sum_i\!\!\sqrt{\sum_j\!\!w_{i,j}^2}$ \cite{l21norm}, is used to drive the units to select a conjunction of descriptive inputs (joints/features) since the $\ell_{21}$ norm can encourage the matrix ${{\bf{W}}_{x\beta,k}}$ to be column sparse. Different groups explore different connection (co-occurrence) patterns in order to acquire the capability of recognizing multiple categories of actions. The $\ell_1$ norm constraint (the second term) helps to learn discriminative joints/features. 

The stochastic gradient descent method is then employed to solve (\ref{equ:opt}). The advantage of the co-occurrence learning is that the model can automatically learn the discriminative joint/feature connections, avoiding the fixed a priori blocking of joint co-occurrences across human parts \cite{CVPR15HRNN} as illustrated in Fig. \ref{fig:Group} (b).
\subsection{3.3 ~~In-depth Dropout for LSTM}
\allowdisplaybreaks
Dropout tries to combine the predictions of many ``thinned'' networks to boost the performance. During training, the network randomly drops some  neurons to force the remaining sub-network to compensate. During testing, the network uses all the neurons together to make predictions.
\begin{figure}[!t]
	\centering
	\begin{subfigure}[t]{0.5\linewidth}
		\centering\includegraphics[width=\textwidth]{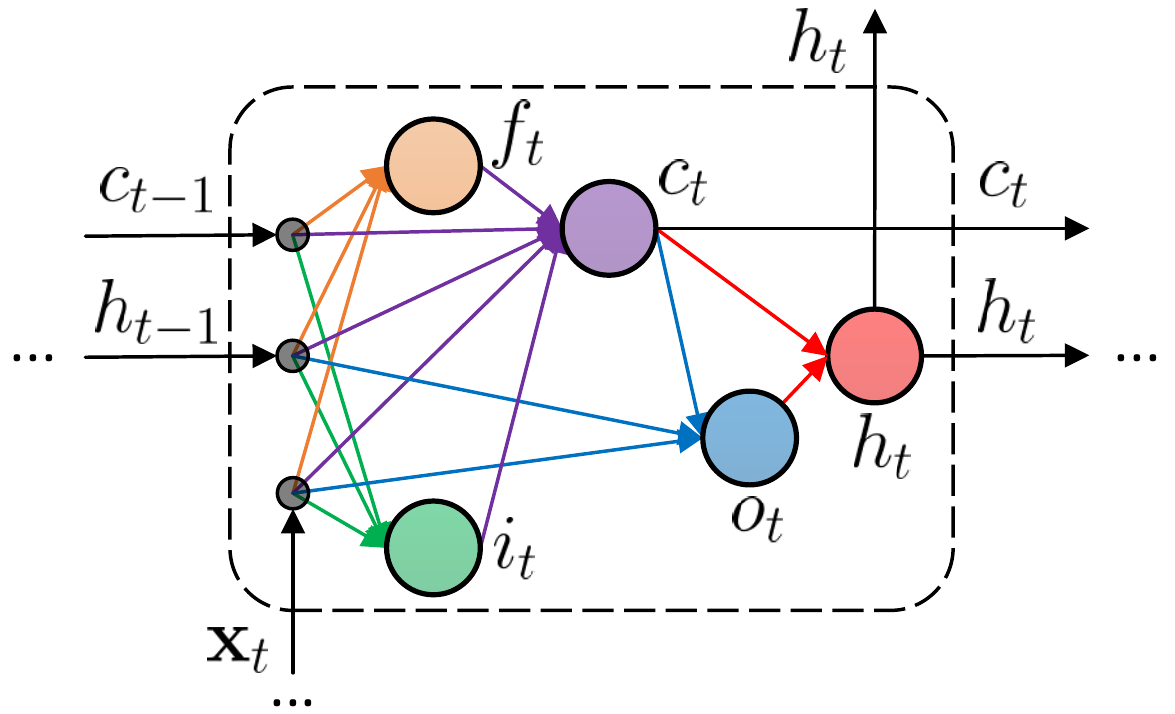}
		\caption{}
		\label{subfig:UnfoldedLSTM}
	\end{subfigure}	
	\begin{subfigure}[t]{0.485\linewidth}
		\centering\includegraphics[width=\textwidth]{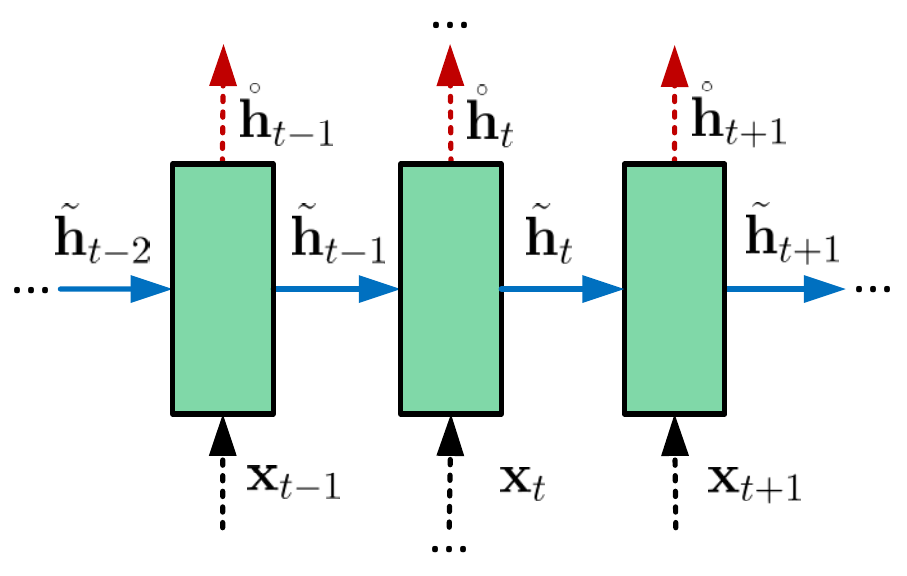}
		\caption{}			
		\label{subfig:DropoutFlow}
	\end{subfigure}
	\caption[]{LSTM dropout. (a) An LSTM neuron viewed in unfolded form. Gates, cell and output response (as marked by large circles) can be dropped. 
		(b) Dropout flow. The solid arrows denote the flow where dropout is forbidden and the dashed arrows denote the flow where dropout is allowed. A rectangle box indicates all the LSTM neurons in this layer.}\label{fig:Dropout}
\end{figure}

To extend this idea to LSTM, we propose a new dropout algorithm to allow the dropping of the internal gates, cell and output response for an LSTM neuron, encouraging each unit to learn better parameters. For clarity, an LSTM neuron is shown in Fig. \ref{fig:Dropout} (a) in the unfolded form, where the units are explicitly connected. For recurrent neural networks, the erasing of all the information from a unit is not expected, especially when the unit remembers events that occurred many timesteps back in the past \cite{ICFHR14DropoutRCNN}. Therefore, we allow the influence of dropout in LSTM to flow along layers (marked by dashed arrows) but prohibit it to flow along the time axis (marked by solid arrows) as illustrated in Fig. \ref{fig:Dropout} (b). To control the influence flows, in the feedforward process, the network calculates and records two types of activations  as follows. The responses of units to be transmitted along the time without dropout are
\begin{equation} 
\label{equ:lstmdropoutrecur}
\begin{aligned}
\!\!\!\!\!\!&\tilde{{\bf{i}}}_t = \sigma \!\left( {{\bf{W}}_{xi}}{\bf{x}}_{t} + {{\bf{W}}_{hi}}\tilde{{\bf{h}}}_{t-1} + {{\bf{W}}_{ci}}\tilde{{\bf{c}}}_{t-1} + {\bf{b}}_i \right), \\
\!\!\!\!\!\!&\tilde{{\bf{f}}}_t = \sigma \!\left( {{\bf{W}}_{xf}}{\bf{x}}_{t} + {{\bf{W}}_{hf}}\tilde{{\bf{h}}}_{t-1} + {{\bf{W}}_{cf}}\tilde{{\bf{c}}}_{t-1} + {\bf{b}}_f \right), \\
\!\!\!\!\!\!&\tilde{{\bf{c}}}_t\!=\tilde{{\bf{f}}}_t \odot \tilde{{\bf{c}}}_{t\!-\!1}+\tilde{{\bf{i}}}_t\odot \tanhmine\!\left( {{\bf{W}}_{xc}}{\bf{x}}_{t}\!+\!{{\bf{W}}_{hc}}\tilde{{\bf{h}}}_{t\!-\!1}\!+\!{\bf{b}}_c \right),\\ 
\!\!\!\!\!\!&\tilde{{\bf{o}}}_t=\sigma \!\left( {{\bf{W}}_{xo}}{{\bf{x}}_{t}} + {{\bf{W}}_{ho}}\tilde{{\bf{h}}}_{t-1} + {{\bf{W}}_{co}}\tilde{{\bf{c}}}_{t} + {\bf{b}}_o \right),\\\!\!\!\!\!\!&\tilde{{\bf{h}}}_t = \tilde{{\bf{o}}}_t \odot \tanhmine \left( \tilde{{\bf{c}}}_t \right).
\end{aligned}
\end{equation} 
The responses of units to be transmitted across layers with dropout applied are
\begin{equation}
\label{equ:lstmdropouthirar}
\begin{aligned}
\!\!\!\!&\drop{{\bf{i}}}_t = \sigma \left( {{\bf{W}}_{xi}}{\bf{x}}_{t} + {{\bf{W}}_{hi}}\tilde{{\bf{h}}}_{t\!-\!1} + {{\bf{W}}_{ci}}\tilde{{\bf{c}}}_{t\!-\!1} + {\bf{b}}_i \right) \!\odot\! {\bf{ m}}_i, \\
\!\!\!\!&\drop{{\bf{f}}}_t = \sigma \left( {{\bf{W}}_{xf}}{{\bf{x}}}_{t}\!+\!{{\bf{W}}_{hf}}\tilde{{\bf{h}}}_{t\!-\!1}\!+\! {{\bf{W}}_{cf}}\tilde{{\bf{c}}}_{t\!-\!1}+{\bf{b}}_f \right) \!\odot\! {\bf{m}}_f,\\
\!\!\!\!&\drop{{\bf{c}}}_t\!=\!\!\left(\!\drop{{\bf{f}}}_t \!\odot\! \tilde{{\bf{c}}}_{t\!-\!1}\!\!+\!\!\drop{{\bf{i}}}_t \!\odot\! \tanhmine\!\left(\! {{\bf{W}}_{xc}}{{\bf{x}}_{t}}\!\!+\!\!{{\bf{W}}_{hc}}\tilde{{\bf{h}}}_{t\!-\!1}\!\!+\!\!{\bf{b}}_c \!\right)\!\right)\! \!\odot\! {\bf{m}}_c,\\
\!\!\!\!&\drop{{\bf{o}}}_t = \sigma \left( {{\bf{W}}_{xo}}{{\bf{x}}_{t}} + {{\bf{W}}_{ho}}\tilde{{\bf{h}}}_{t-1} + {{\bf{W}}_{co}}\drop{{\bf{c}}}_{t} + {\bf{b}}_o \right) \!\odot\! {\bf{m}}_o, \\
\!\!\!\!&\drop{{\bf{h}}}_t = \drop{{\bf{o}}}_t \odot \tanhmine \left( \drop{{\bf{c}}}_t \right) \odot {\bf{m}}_h,
\end{aligned}
\end{equation}
where ${\bf{m}}_i, {\bf{m}}_f, {\bf{m}}_c, {\bf{m}}_o,$ and ${\bf{m}}_h$ are dropout binary mask vectors for input gates, forget gates, cells, output gates and output responses, respectively, with an element value of 0 indicating that dropout happens. Note that for the first LSTM layer, the inputs ${\bf{x}}_{t}$ are the skeleton joints of a frame; for the higher LSTM layer, the inputs ${\bf{x}}_{t}$ are the response outputs of the previous layer. 

During the training process, the errors back-propagated to the output responses ${\bf{h}}_{t}$ are
\begin{eqnarray}
\begin{aligned}
&{\bm{\epsilon}}_{h}^t = {\drop{\bm\epsilon}}_{h}^t + {\tilde{\bm\epsilon}}_{h}^t,\\
&{\drop{\bm\epsilon}}_{h}^t = {\bm\epsilon}_{h}^{hier} \odot {\bf{m}}_h, 
&{\tilde{\bm\epsilon}}_{h}^t = {\bm\epsilon}_{h}^{recur},
\end{aligned}
\end{eqnarray}
where ${\bm\epsilon}_{h}^{hier}$ denotes the vector of errors back-propagated from the upper layer at the same time slot,  ${\bm\epsilon}_{h}^{recur}$ denotes the vector of errors from the next time slot in the same layer, $
\drop{\bm\epsilon}^t$ and $\tilde{\bm\epsilon}^t$ denote the dropout errors from the upper layer and recurrent errors from the next time slot, respectively.

By taking derivative of $\drop{{\bf{h}}}_t$ with respect to $\drop{{\bf{o}}}_t$ based on (\ref{equ:lstmdropouthirar}), we get the errors from $\drop{{\bf{h}}}_t$ to $\drop{{\bf{o}}}_t$ which represent the errors from upper layer with dropout involved 
\begin{equation}
\begin{aligned}
&{\drop{\bm\epsilon}}_{o}^t = (\drop{{\bm\epsilon}}_{h}^t \odot\frac{\partial \drop{{\bf{h}}}_t}{\partial \drop{{\bf{o}}}_t})\odot {\bf{m}}_o = \drop{{\bm\epsilon}}_{h}^t \odot \tanhmine \left( \drop{{\bf{c}}}_t \right)\odot {\bf{m}}_o.
\end{aligned}
\end{equation}
Similarly, based on (\ref{equ:lstmdropoutrecur}), we get the errors back-propagated from $\tilde{{\bf{h}}}_t$ to $\tilde{{\bf{o}}}_t$ which represent the errors from the next time slot in the same layer without dropout
\begin{equation}
\begin{aligned}
&{\tilde{\bm\epsilon}}_{o}^t = \tilde{{\bm\epsilon}}_{h}^t \odot \frac{\partial \tilde{{\bf{h}}}_t}{\partial \tilde{{\bf{o}}}_t} = \tilde{{\bm\epsilon}}_{h}^t \odot \tanhmine \left( \tilde{{\bf{c}}}_t \right).
\end{aligned}
\end{equation}
Then, the errors back-propagated to the output gates are the summation of the two types of errors
\begin{equation}
\begin{aligned}
&{\bm\epsilon}_{o}^t = {\drop{\bm\epsilon}}_{o}^t + {\tilde{\bm\epsilon}}_{o}^t.
\end{aligned}
\end{equation}
In the same way, we derive the errors propagated to the cells, forget gates, and input gates, based on (\ref{equ:lstmdropoutrecur}) and (\ref{equ:lstmdropouthirar}).

During the testing process, we use all the neurons but multiplying the units of LSTM neurons (in the LSTM layer where dropout is applied) by the probability values of $1-p$, where $p$ is the dropout probability of that unit. Note that the simple dropout which only drops the output responses ${\bf{h}}_{t}$ \cite{ICLR15DropoutLSTM} is a special case of our proposed dropout. 

\subsection{3.4 ~~Action Recognition using the Learned Model}
With the learned deep LSTM network, the probability that a sequence ${\bf{X}}$ belongs to the class $C_k$ is
\begin{eqnarray}
\begin{aligned}
&p(C_k|{\bf{X}}) = \frac{e^{o_k}}{\sum_{i=1}^{C} e^{o_i}},~~ k = 1, \cdots, C, \\
&{\bf{o}} = \sum_{t=1}^{T} \left( {{\bf{W}}_{\overrightarrow{h}o} \overrightarrow{{\bf{h}}}_{t} + {\bf{W}}_{\overleftarrow{h}o} \overleftarrow{{\bf{h}}}_{t} + {\bf{b}}_o} \right),
\end{aligned}
\end{eqnarray}
where $C$ denotes the number of classes, $T$ represents the length of the test sequence, ${\bf{o}} = [o_1, o_2,\cdots,o_C]$, $\overrightarrow{{\bf{h}}}_{t}$ and $\overleftarrow{{\bf{h}}}_{t}$ denote the output responses of the last bidirectional LSTM layer. Then, the class with the highest probability is chosen as action class. 

%

\section{4 ~~Experiments}
We validate the proposed model on the SBU kinect interaction dataset \cite{CVPRW12TwoPerson}, HDM05 dataset \cite{HDM05}, and CMU dataset \cite{MoCap} whose groundtruth was labeled by ourselves. We have also tested our model on the Berkeley MHAD action recognition dataset \cite{MHAD} and achieved 100\% accuracy. To investigate the impact of each component in our model, we conduct experiments under different configurations represented as follows:
\begin{itemize}
\item Deep LSTM is our basic scheme without regularizations;
\item Deep LSTM + Co-occurrence is the scheme with our proposed co-occurrence regularization applied;
\item Deep LSTM + Simple Dropout is the scheme with the dropout algorithm proposed by Zaremba et al. \cite{ICLR15DropoutLSTM}  applied to our basic scheme; 
\item Deep LSTM + In-depth Dropout is the scheme with our proposed in-depth dropout applied;
\item Deep LSTM + Co-occurrence + In-depth Dropout is our final scheme with both co-occurrence regularization and in-depth dropout applied.
\end{itemize}


Down-sampling the skeleton sequences in temporal is performed to have the frame rate of 30fps on the HDM05 dataset and CMU dataset. To reduce the influence of noise in the captured skeleton data, we smooth each joint's position of the raw skeleton using the filter $\left(-3, 12, 17, 12, -3\right)/35$ in the temporal domain \cite{smooth,CVPR15HRNN}. The number of groups ($K$) in our model is set to 5, 10, and 10 for the first three layers experimentally. We set the dropout probability $p$ to 0.2 for each unit in an LSTM neuron, which makes the overall dropout probability of an LSTM neuron approach 0.5 (this can be derived based on (\ref{equ:lstmdropouthirar})). Note that when dropout is applied, the number of neurons in the corresponding layer is doubled as suggested by previous work \cite{JMLR14Dropout}. We set the parameters $\lambda_1$ and $\lambda_2$ in (\ref{equ:opt}) experimentally. 

\subsection{4.1 ~~SBU Kinect Interaction Dataset}

The SBU kinect interaction dataset is a Kinect captured human activity recognition dataset depicting two person interaction, which contains 230 sequences of 8 classes (6,614 frames) with subject independent 5-fold cross validation. The smoothed positions of joints are used as the input of the deep LSTM network for recognition. The number of neurons is set to 100$\times 2$, 100, 110$\times 2$, 110, 200$\times 2$ for the first to fifth layers respectively, where 2 indicates bidirectional LSTM is used and thus the number of neurons is doubled. 

We have compared our schemes with other skeleton based methods \cite{CVPRW12TwoPerson,ICMEWHumanInteraction,CVPR15HRNN}. Note that we add an additional layer to fuse the two subnets corresponding to the two persons when extending Hierarchical RNN scheme for use in the two person interaction scenario \cite{CVPR15HRNN}. We summarize the results in terms of the average recognition accuracy (5-fold cross validation) in Table \ref{tab:sbu}. 
\begin{table}[ht]
	\fontsize{9pt}{10pt}\selectfont\centering
	\caption{Comparisons on SBU kinect interaction dataset.}\label{tab:sbu}
	\begin{tabular}{c|c}
		\hlinew{0.9pt}
		Methods&Acc.(\%)\\		
		\hlinew{0.7pt}
		\tabincell{c}{Raw skeleton \cite{CVPRW12TwoPerson}}&49.7\\
		\hline \tabincell{c}{Joint feature \cite{CVPRW12TwoPerson}}&80.3\\
		\hline \tabincell{c}{Raw skeleton \cite{ICMEWHumanInteraction}}&79.4\\
		\hline \tabincell{c}{Joint feature \cite{ICMEWHumanInteraction}}&86.9\\
		\hline \tabincell{c}{Hierarchical RNN \cite{CVPR15HRNN}} &80.35\\
		\hlinew{0.9pt} Deep LSTM & 86.03\\
		\hline \tabincell{c}{Deep LSTM + Co-occurrence} & 89.44\\
		\hline Deep LSTM + Simple Dropout & 89.70\\
		\hline Deep LSTM + In-depth Dropout& 90.10\\
		\hline
		\tabincell{c}{Deep LSTM+Co-occurrence+In-depth Dropout} & \textbf{90.41}\\
		\hlinew{0.9pt}
	\end{tabular}
\end{table}

Table \ref{tab:sbu} shows that our basic scheme of Deep LSTM achieves comparable performance to the method using handcrafted complex features \cite{ICMEWHumanInteraction}. The proposed schemes of Deep LSTM + Co-occurrence and Deep LSTM + In-depth Dropout can improve the recognition accuracy by 3.4\% and 4.1\% respectively over Deep LSTM, indicating that the co-occurrence exploration boosts the discrimination of features and the proposed LSTM dropout is capable of learning a more effective model. Deep LSTM + In-depth Dropout is superior to Deep LSTM + Simple Dropout. Note that the deep LSTM network achieves remarkable (5.6\%) performance improvement in comparison with the hierarchical RNN network \cite{CVPR15HRNN}. That is because allowing full connection of joints/features with neurons rather than imposing a priori subnet constraints facilitates the interaction among joints especially when the joints do not belong to the same part, or same person. Our scheme with combined regularizations achieves the best performance.

\subsection{4.2 ~~HDM05 Dataset}
The HDM05 dataset contains 2,337 skeleton  sequences performed by 5 actors (184,046 frames after down-sampling). For fair comparison, we use the same protocol (65 classes, 10-fold cross validation) as used by Cho and Chen \cite{VISAPP14Classify}. The pre-processing is the same as that done in the hierarchical RNN scheme \cite{CVPR15HRNN} (centralize the joints' positions to human center for each frame and smooth the positions). The number of neurons is 100$\times 2$, 110, 120$\times 2$, 120, 200$\times 2$ for the five layers respectively. Table \ref{tab:hdm05} shows the results in terms of average accuracy. Our basic deep LSTM achieves better results than the Multi-layer Perception model, which suggests that LSTM exhibits better motion modeling ability than the MLP. With the proposed co-occurrence learning and in-depth dropout regularization, our full model also performs better than the manually designed 
hierarchical RNN approach. 
\begin{table}[ht]
	\fontsize{9pt}{10pt}\selectfont\centering
	\caption{Comparisons on HDM05 dataset.}\label{tab:hdm05}
	\begin{tabular}{c|c}
		\hlinew{0.9pt}
		Methods&Acc.(\%)\\
		\hlinew{0.7pt} \tabincell{c}{Multi-layer Perceptron \cite{VISAPP14Classify}}&95.59\\
		\hline \tabincell{c}{Hierarchical RNN \cite{CVPR15HRNN}} &96.92\\
		\hlinew{0.9pt} Deep LSTM & 96.80\\
		\hline \tabincell{c}{Deep LSTM + Co-occurrence} & 97.03\\
		\hline Deep LSTM + Simple Dropout & 97.21\\
		\hline Deep LSTM + In-depth Dropout& 97.25\\
		\hline \tabincell{c}{Deep LSTM+Co-occurrence+In-depth Dropout} & \textbf{97.25}\\
		\hlinew{0.9pt}
	\end{tabular}
\end{table}

\subsection{4.3 ~~CMU Dataset}


We have categorized the CMU motion capture dataset into 45 classes for the purpose of skeleton based action recognition\footnote{\label{foot1}http://www.escience.cn/people/zhuwentao/29634.html}. The categorized dataset contains 2,235 sequences
(987,341 frames after down-sampling) and is the largest
skeleton based human action dataset so far. This dataset
is much more challenging because: (i) the lengths of sequences vary greatly; (ii) the within-class diversity is large, e.g., for ``walking", different people walk at different speeds and along different paths; (iii) the dataset contains complex actions such as dance, doing yoga.

We have evaluated the performance on both the entire dataset (CMU) and a subset of the dataset (CMU subset). For this subset, we have chosen eight representative action categories containing 664 sequences (125,667 frames after down-sampling), with actions of \emph{jump}, \emph{walk back}, \emph{run}, \emph{sit}, \emph{getup}, \emph{pickup}, \emph{basketball}, \emph{cartwheel}. The same pre-processing as used for the HDM05 dataset is performed. The number of neurons is set to 100$\times 2$, 100, 120$\times 2$, 120, 100$\times 2$ for the five layers. Three-fold cross validation is conducted and the results in terms of average accuracy are shown in Table \ref{tab:cmu}. We can see that the proposed model achieves significant performance improvement, indicating that it can better learn the discriminative features and model long-range temporal dynamics even for this challenging dataset. 

\begin{table}[ht]
	\fontsize{9pt}{10pt}\selectfont\centering
	\caption{Accuracy (\%) comparisons on CMU dataset.}\label{tab:cmu}
	\begin{tabular}{c|c|c}
		\hlinew{0.9pt} Methods&CMU subset&CMU \\
		\hlinew{0.7pt} \tabincell{c}{Hierarchical RNN \\ \cite{CVPR15HRNN}}&83.13&75.02\\
		\hlinew{0.9pt} Deep LSTM & 86.00&79.53\\  
				\hline \tabincell{c}{Deep LSTM + Co-occurrence} & 87.20&79.60\\
	    \hline Deep LSTM + Simple Dropout & 87.80&80.59\\
		\hline Deep LSTM + In-depth Dropout& 88.25&80.99\\
		\hline \tabincell{c}{Deep LSTM+\\Co-occurrence+In-depth Dropout} & \textbf{88.40}&\textbf{81.04}\\ 
		\hlinew{0.9pt}
	\end{tabular}
\end{table}
\subsection{4.4 ~~Discussions}\label{ssec:discussion}
\begin{figure}[ht]
	\centering
	\begin{subfigure}[t]{0.495\linewidth}
		\centering\includegraphics[width=\linewidth]{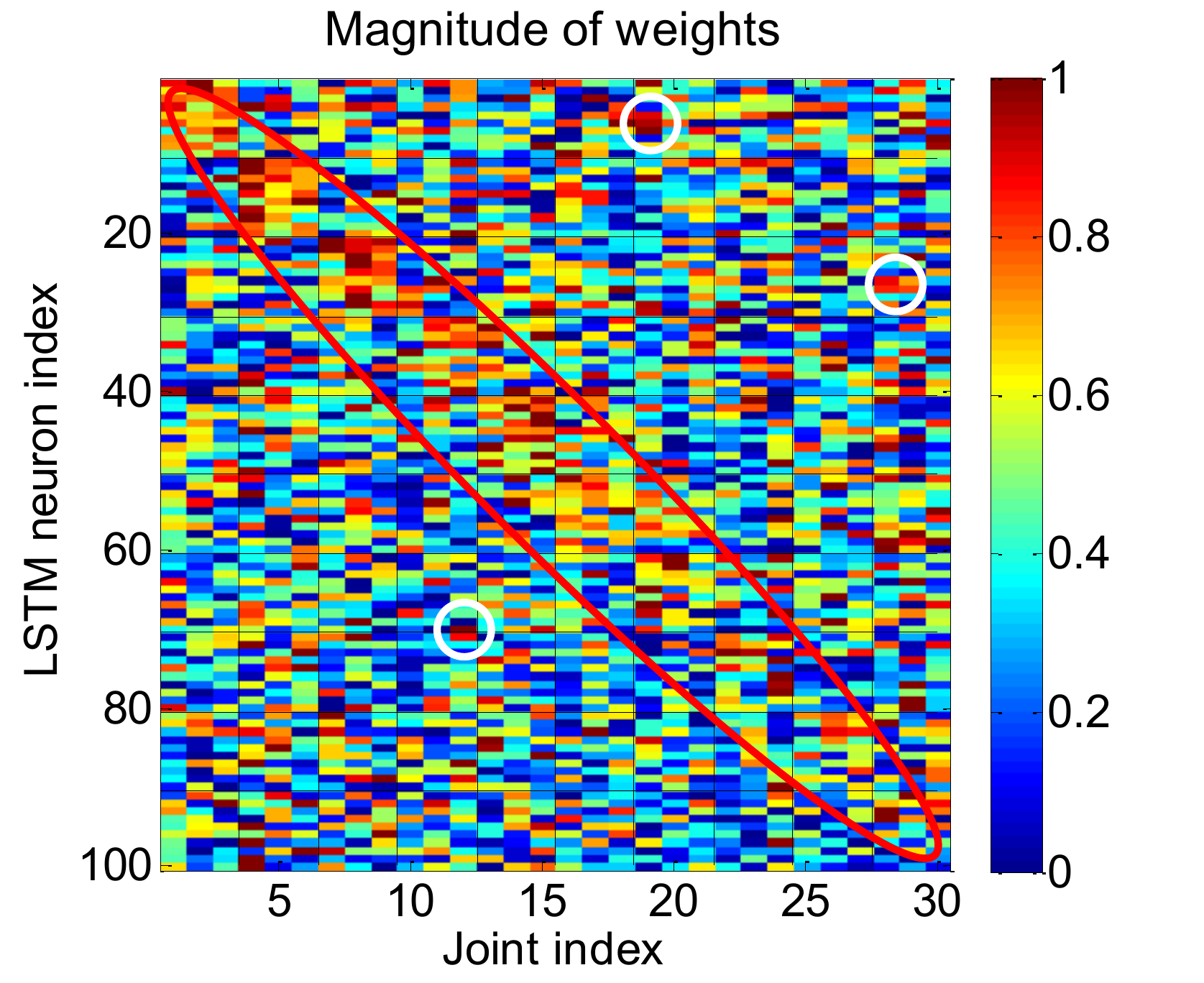} 
		\caption{}
		\label{subfig:weightl1}
	\end{subfigure}
	\begin{subfigure}[t]{0.495\linewidth}
		\centering\includegraphics[width=\linewidth]{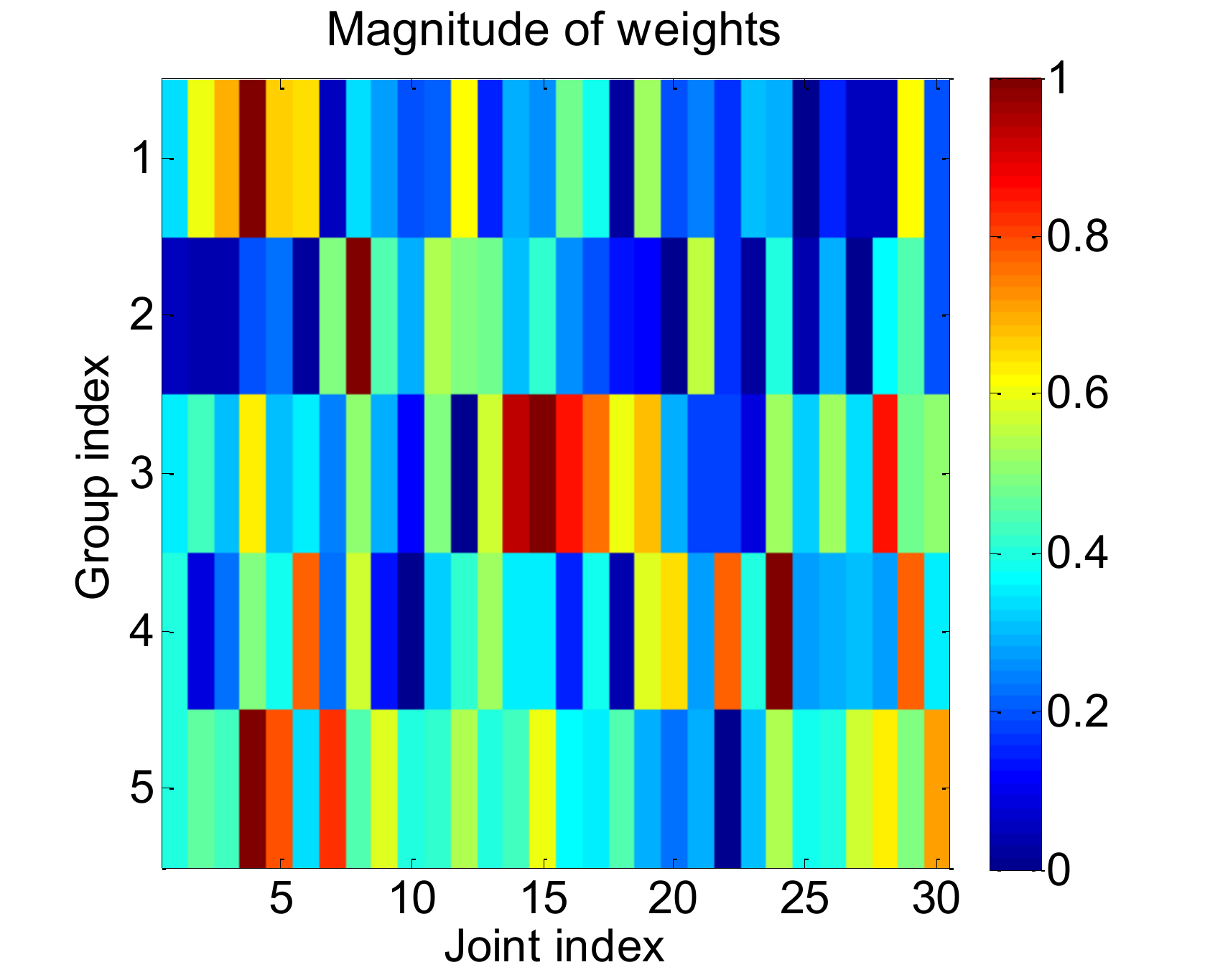}
		\caption{} 
		\label{subfig:weightgroup}
	\end{subfigure}
	\caption[]{Visualization of the absolute values of input gate weights in the first layer on SBU kinect interaction dataset. Horizontal axis denotes the indexes of 30 joints of two persons. (a) Vertical axis denotes the 100 LSTM neurons. Each element represents the absolute value of the weight between the corresponding joint and input gate unit of that LSTM neuron. Every three nearby joints form a part of a person. (b) Vertical axis denotes the five groups of LSTM neurons. We observe different groups have different weight patterns, preferring different conjunctions of joints.}\label{fig:weightvisual}
\end{figure}

To further understand our deep LSTM network, we visualize the weights learned in the first LSTM layer on the SBU kinect interaction dataset in Fig. \ref{fig:weightvisual} (a). Each element represents the absolute value of the weight between the corresponding skeleton joint and input gate of that LSTM neuron. It is observed that the weights in the diagonal positions marked by the red ellipse have high values, which means the co-occurrence regularization helps learn the human parts automatically. In contrast to the part based subnet fusion model \cite{CVPR15HRNN}, the learned co-occurrences of joints by our model do not limit the connections to be in the same part, as there are many large weights outside the diagonal regions, e.g., in the regions marked by white circles, making the network more powerful for action recognition. This also signifies the importance of the proposed full connection architecture. By averaging the energy of the weights in the same group of neurons for each joint, we obtain Fig. \ref{fig:weightvisual} (b) which has five groups of LSTM neurons. It is observed that different groups have different weight patterns, preferring different conjunctions of joints.

\section{5 ~~Conclusion}
\label{sec_conclusion}
\noindent In this paper, we propose an end-to-end fully connected deep LSTM network for skeleton based action recognition. The proposed model facilitates the automatic learning of feature co-occurrences from the skeleton joints through our designed regularization. To ensure effective learning of the deep model, we design an in-depth dropout algorithm for the LSTM neurons, which performs dropout for the internal gates, cell, and output response of the LSTM neuron. Experimental results demonstrate the state-of-the-art performance of our model on several datasets.

\section{Acknowledgment}
We would like to thank David Wipf from Microsoft Research Asia for the valuable discussions, and thank Yong Du from Institute of Automation, Chinese Academy of Sciences for providing Hierarchical RNN code for the comparison.

\bibliographystyle{aaai}
\small
\bibliography{AAAI15Skeleton}

\end{document}